\documentclass[conference,a4paper]{IEEEtran}
\usepackage{cite}

\usepackage{times}
\usepackage{epsfig}
\usepackage{graphicx}
\usepackage[cmex10]{amsmath}
\usepackage{amssymb}
\usepackage{array}
\usepackage{ragged2e}
\usepackage{mdwmath}
\usepackage{mdwtab}
\usepackage{algorithm}
\usepackage{algpseudocode}
\algdef{SE}[DOWHILE]{Do}{doWhile}{\algorithmicdo}[1]{\algorithmicwhile\ #1}%
\algnewcommand\algorithmicinput{\textbf{Input:}}
\algnewcommand\algorithmicoutput{\textbf{Output:}}
\algnewcommand\Input{\item[\algorithmicinput]}
\algnewcommand\Output{\item[\algorithmicoutput]}
\usepackage{multirow}
\ifCLASSINFOpdf
\else
\fi
\begin{document}
\title{Universal Wavelet Units in 3D Retinal Layer Segmentation}

\author{\IEEEauthorblockN{An D. Le\IEEEauthorrefmark{1}, Hung Nguyen\IEEEauthorrefmark{1}, Melanie Tran\IEEEauthorrefmark{2}, Jesse Most\IEEEauthorrefmark{2},\\ Dirk-Uwe G. Bartsch\IEEEauthorrefmark{2}, William R Freeman\IEEEauthorrefmark{2}, Shyamanga Borooah\IEEEauthorrefmark{2}, Truong Q. Nguyen\IEEEauthorrefmark{1}, and Cheolhong An\IEEEauthorrefmark{1}}

\IEEEauthorblockA{\IEEEauthorrefmark{1}Jacobs School of Engineering, University of California San Diego, La Jolla, CA 92093, USA \\ 
\{d0le,hun004,tqn001,chan\}@ucsd.edu\\
\IEEEauthorrefmark{2}Jacobs Retina Center, Shiley Eye Institute, University of California San Diego, La Jolla, CA 92093, USA\\
\{mdtran,jmost,dbartsch,wrfreeman,sborooah\}@health.ucsd.edu
}}


\maketitle

\begin{abstract}
%\boldmath
This paper presents the first study to apply tunable wavelet units (UwUs) for 3D retinal layer segmentation from Optical Coherence Tomography (OCT) volumes. To overcome the limitations of conventional max-pooling, we integrate three wavelet-based downsampling modules—OrthLattUwU, BiorthLattUwU, and LS-BiorthLattUwU—into a motion-corrected MGU-Net architecture. These modules use learnable lattice filter banks to preserve both low- and high-frequency features, enhancing spatial detail and structural consistency. Evaluated on the Jacobs Retina Center (JRC) OCT dataset, our framework shows significant improvement in accuracy and Dice score, particularly with LS-BiorthLattUwU, highlighting the benefits of tunable wavelet filters in volumetric medical image segmentation.
\end{abstract}


% \fi
\IEEEpeerreviewmaketitle
\section{Introduction}
Optical Coherence Tomography (OCT) enables high-resolution imaging of retinal structures and is essential for diagnosing diseases like age-related macular degeneration (AMD), diabetes-related macular edema (DME), and glaucoma. Accurate retinal layer segmentation supports disease monitoring but remains challenging due to motion artifacts and the limitations of 2D deep learning methods that ignore 3D context.
To address these issues, we enhance the MGU-Net architecture based model \cite{Yiqian1,Yiqian2,Yiqian3} by replacing max-pooling layers with wavelet-inspired downsampling modules: OrthLattUwU\cite{OrthLatt_UwU}, BiorthLattUwU\cite{BiorLattUwU}, and the proposed LS-BiorthLattUwU. These modules use tunable lattice filter banks to better preserve spatial details and high-frequency features. We further incorporate an attention head to adaptively integrate subband information. Applied to 3D OCT segmentation with motion correction, our UwU-MGUNet framework showed improved performance based on accuracy and Dice scores, especially with LS-BiorthLattUwU. Experiments on the JRC dataset demonstrate superior performance over the standard pooling-based model.
\section{Related Works}
Early OCT segmentation approaches, such as level-set models \cite{carass2014multiple,novosel2015loosely} and graph-based methods \cite{chiu2010automatic,rathke2014probabilistic}, faced limitations in robustness and scalability, particularly in pathological cases. With the rise of deep learning, models like U-Net \cite{ronneberger2015u}, RelayNet \cite{roy2017relaynet}, and cascaded FCNs \cite{he2017towards} significantly improved performance by learning pixel-wise segmentation. However, most operate on 2D B-scans and neglect 3D spatial coherence. To address 3D consistency, architectures like MGU-Net \cite{li2021multi} incorporated graph-convolution-based reasoning, and joint motion correction with 3D segmentation networks were introduced \cite{Yiqian2}, improving alignment across B-scans. Yet, conventional max-pooling used in these models often discards fine spatial detail.
Wavelet-based pooling techniques \cite{AnLe, OrthLatt_UwU, BiorLattUwU} have shown promising performance by preserving high-frequency features via learnable lattice filter banks in clinical applications \cite{wavelet_rp, rp_clinical, uwu_ILM-ERM, ilm-erm_clinical}. These methods, especially when integrated with attention and channel-wise fusion mechanisms \cite{chen2019graph}, enhance feature selectivity and global reasoning—proving effective in classification and anomaly detection. However, the application of such wavelet-inspired downsampling in 3D retinal OCT segmentation remains limited. Our work aims to bridge this gap by integrating OrthLattUwU, BiorthLattUwU, and the proposed method, LS-BiorthLattUwU, into a 3D segmentation framework, improving layer delineation and spatial consistency in diseased retinas.
\section{Method}
\subsection{OrthLatt-UwU: Tunable Orthogonal Wavelet Filters with Lattice Structure}
OrthLatt-UwU\cite{OrthLatt_UwU} is a unit that leverages both low-frequency and high-frequency components from the DWT analysis to find the optimal feature map. Instead of utilizing predefined wavelets, OrthLatt-UwU, characterized by its trainable coefficients, is constructed using a lattice structure. In addition, the perfect reconstruction characteristic of DWT can be achieved through analysis and synthesis parts of a filter bank. For the analysis component, $\textbf{H}_{0}$ and $\textbf{H}_{1}$ are low-pass and high-pass filters, correspondingly. Conversely, $\textbf{F}_{0}$ and $\textbf{F}_{1}$ are low-pass and high-pass filters for the synthesis part, respectively.
To achieve perfect reconstruction, the aliasing cancellation condition must be fulfilled and there should be no distortion in the reconstructed signal \cite{Mallat,Wavelets_and_filter_banks}. To satisfy the alias cancellation condition, given $\textbf{h}_{0}=\textbf{[}h(0), h(1),..., h(N-1)\textbf{]}$ as coefficients of $\textbf{H}_{0}$ with $N$ taps, the coefficients of the other filters in the orthogonal filter bank can be deduced through sign alternating flip, order flip and alternating signs relations \cite{Wavelets_and_filter_banks}, which can be expressed as follows:
\begin{equation}
\label{eq:AliasingCancellation9}
\begin{cases} 
\mbox{{\footnotesize \textbf{Order Flip:}}} & {\small \textbf{f}_{0}(n)=\textbf{h}_{0}(N-1-n)} \\
\mbox{{\footnotesize \textbf{Sign Alternating Flip:}}} & {\small \textbf{h}_{1}(n)=(-1)^{n}\textbf{h}_{0}(N-1-n)} \\
\mbox{{\footnotesize \textbf{Alternating Sign:}}} &{\small \textbf{f}_{1}(n)=-(-1)^{n}\textbf{h}_{0}(n),}
\end{cases} 
\end{equation}
where $\textbf{f}_{0}$, $\textbf{h}_{1}$, and $\textbf{f}_{1}$ are filter coefficients of $\textbf{F}_{0}$, $\textbf{H}_{1}$, and $\textbf{F}_{1}$, respectively. From the relations presented in Eq. \eqref{eq:AliasingCancellation9}, the filter bank satisfies the anti-aliasing condition. Moreover, with the aliasing cancellation condition, filter coefficients $\textbf{h}_{0}$ of $\textbf{H}_{0}$ is designed, which reduces the number of parameters needed for the analysis part used in a classification model. Then, in order to find a $\textbf{H}_{0}$ that ensures no distortion in the reconstructed signal, one approach is to impose the orthogonal structure in the filter by building it with lattice blocks \cite{Wavelets_and_filter_banks}, which can be expressed as follows:
\begin{multline}
\label{eq:latticestructure9}
 \begin{bmatrix}
   \textbf{H}_{0}(z) \\
   \textbf{H}_{1}(z) \\
   \end{bmatrix} = \begin{bmatrix}
   \textbf{H}_{0}(z) \\
   -z^{-(N-1)}\textbf{H}_{0}(-z^{-1}) \\
   \end{bmatrix} \\
   = \begin{bmatrix} 1&0 \\0&-1 \\ \end{bmatrix}\textbf{R}_{K}\Lambda(z^2)\cdots\textbf{R}_{1}\Lambda(z^2)\textbf{R}_{0}
   \begin{bmatrix}
   1 \\
   z^{-1} \\
 \end{bmatrix},
\end{multline}
where $\textbf{R}_{k}$ is a rotation matrix constructing the filter with $k = 0,\cdots,K$. The delay matrices within the filter are represented by $\textbf{$\Lambda$}(z^2)$. In addition, $N$ is the order of the filter which can be defined as $N = 2K + 1$. In this work, to ensure the half-band condition, we use rotation matrices, which inherently are orthogonal matrices. Hence, $\textbf{R}_{k}$ and $\textbf{$\Lambda$}(z)$ can be mathematically expressed as follows:
\begin{gather}
\label{eq:RotationMatrix9}
 \textbf{R}_{k} = \begin{bmatrix} cos(\theta_{k})&sin(\theta_{k}) \\-sin(\theta_{k})&cos(\theta_{k}) \\ \end{bmatrix} = \begin{bmatrix} c_k&s_k \\-s_k&c_k \\ \end{bmatrix}.
\end{gather}
\begin{gather}
\label{eq:DelayMatrix9}
 \textbf{$\Lambda$}(z) = \begin{bmatrix} 1&0 \\0&z^{-1} \\ \end{bmatrix}.
\end{gather}
In Eq. \eqref{eq:RotationMatrix9}, $\theta_{k}$ is a rotation angle determining the coefficients of the wavelet filter bank with $k = 0,...,K$. These rotation angles in the rotation matrices, where either their rows or columns are orthonormal to each other, are also termed lattice coefficients and determine the coefficients of the filterbank's filters. This orthonormality consequently ensures that the filters, which result from the multiplication of rotation and delay matrices, maintain orthogonality.
Instead of training the filter coefficients directly, our method integrates these lattice coefficients as trainable parameters for the wavelet unit. With lattice coefficients, the filter bank is structural orthogonal and thus satisfies perfect reconstruction. Moreover, the number of lattice coefficients is ($K+1$), which is approximately half the number of filter coefficients $N$. 
\subsection{BiorthLatt-UwU: Tunable Biorthogonal Wavelet Filters with Lattice Structure}

Biorthogonal-Lattice TypeA UwU (BiorthLatt-UwU) leverages both low-frequency and high-frequency components from a decomposition process (analysis) of filter-banks with biorthogonal lattice structure to find the optimal feature map \cite{BiorLattUwU}. In addition, as biorthogonal lattice structures impose the perfect reconstruction characteristic, the original signal can be reconstructed, which is called synthesis process. For the analysis component, ${H}_{0}(z)$ and ${H}_{1}(z)$ are finite impulse response (FIR) analysis filters while ${F}_{0}(z)$ and ${F}_{1}(z)$ are FIR synthesis filters. The filter banks can also be represented with poly-phase representation. The relationship between the poly-phase matrices and the filters ${H}_{0}(z)$, ${H}_{1}(z)$, ${F}_{0}(z)$ and ${F}_{1}(z)$ can be mathematically represented as follows:

\begin{equation}
\label{eq:polyphase_filter_9}
\begin{cases} 
\begin{bmatrix} 
    {H}_{0}(z) \\
    {H}_{1}(z) \\
   \end{bmatrix} = \textbf{E}(z^{2}) \begin{bmatrix}
   1 \\
   z^{-1} \\
   \end{bmatrix}\\
   \\
\begin{bmatrix} 
    {F}_{0}(z) \quad
    {F}_{1}(z) \\
   \end{bmatrix} =  \begin{bmatrix}
   z^{-1}\quad 1\end{bmatrix} \textbf{R}(z^{2})\\
\end{cases}
\end{equation}
In order to achieve perfect reconstruction (PR), we should have $\textbf{R}(z) = \textbf{E}^{-1}(z)$. Therefore, the relationship between the analysis and synthesis filters of a FIR PR filter bank can be mathematically illustrated as follows:
\begin{equation}
\label{eq:polyphase_filter2_9}
\begin{bmatrix} 
    {F}_{0}(z) \\
    {F}_{1}(z) \\
   \end{bmatrix} = \begin{bmatrix}
   -{H}_{0}(-z) \\
   {H}_{1}(-z) \\
   \end{bmatrix}.\\
\end{equation}

\begin{figure}[t!]
\begin{center}
\includegraphics[width=\linewidth]{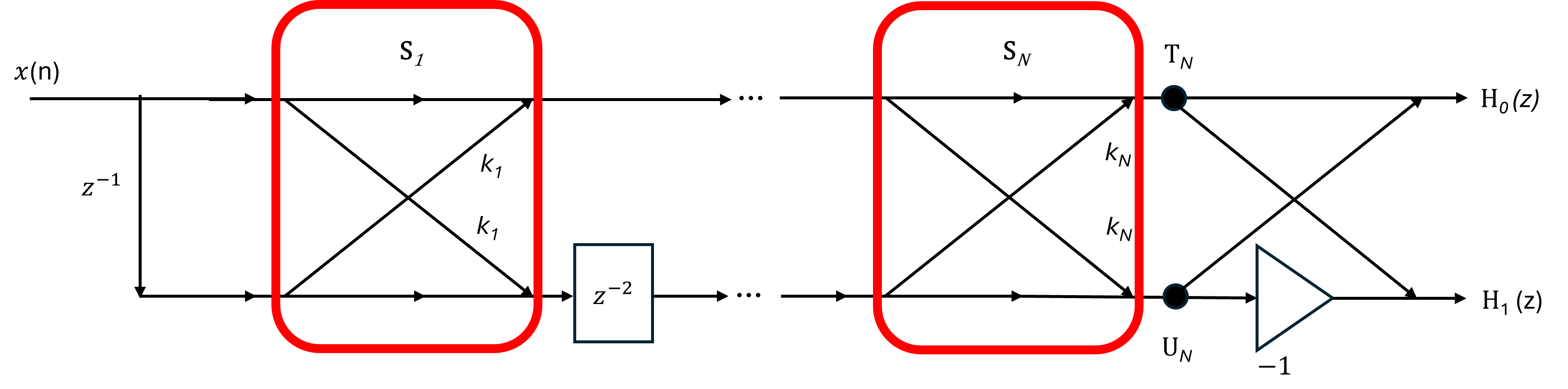}
\end{center}
\caption{\justifying{Analysis filter-bank with biorthogonal lattice structure diagram. The diagram shows the ${H}_{0}(z)$ and ${H}_{1}(z)$ with order N can be represented as a multiplication of multiple biorthogonal lattice matrices $\textbf{S}_{m}$ with lattice coefficient $k_{m}$ for $m$ ranging from 1 to $N$.}} 
\label{fig:bior-latt-TypeA-AnalysisFilterBank_9}
\end{figure}

With Eq. \eqref{eq:polyphase_filter2_9}, synthesis filters can be found given the established analysis filters. In this work, we build our BiorLatt-UwU with biorthogonal lattice structure by having a linear-phase FIR filter pair. The biorthogonal filters to be designed will have a symmetric ${H}_{0}(z)$ and an asymmetric ${H}_{1}(z)$. We call this biorthogonal lattice structure type A as there is also another biorthogonal lattice  with a symmetric ${H}_{0}(z)$ and ${H}_{1}(z)$ \cite{BiorLattictStructure}. In this work, to simplify the design and implementation in the CNN architecture, we assume that ${H}_{0}(z)$ and ${H}_{1}(z)$ have the same odd order. With the type A biorthogonal lattice structure, the ${H}_{0}(z)$ and ${H}_{1}(z)$ pair can be implemented as a cascade of multiple biorthogonal lattice matrices, which can be expressed as follows:
\begin{align}
\label{eq:biorlatticestructure_9}
 \begin{bmatrix}
   {H}_{0}(z) \\
   {H}_{1}(z) \\
   \end{bmatrix}
   &= \begin{bmatrix} 1&1 \\1&-1 \\ \end{bmatrix}\begin{bmatrix}
   {T}_{N}(z) \\
   {U}_{N}(z) \\
   \end{bmatrix} \\
   &= \begin{bmatrix} 1&1 \\1&-1 \\ \end{bmatrix} \begin{bmatrix} 1&k_{N} \\k_{N}&1 \\ \end{bmatrix} \begin{bmatrix} 1&0 \\0&z^{-2} \\ \end{bmatrix} \cdots \nonumber \\
  &\cdots\begin{bmatrix} 1&k_{2} \\k_{2}&1 \\ \end{bmatrix} \begin{bmatrix} 1&0 \\0&z^{-2} \\ \end{bmatrix} \begin{bmatrix} 1&k_{1} \\k_{1}&1 \\ \end{bmatrix}
   \begin{bmatrix}
   1 \\
   z^{-1} \\
 \end{bmatrix}\\
   &= \begin{bmatrix} 1&1 \\1&-1 \\ \end{bmatrix}\textbf{S}_{N}\Lambda(z^{2})\cdots\textbf{S}_{2}\Lambda(z^{2})\textbf{S}_{1}
   \begin{bmatrix}
   1 \\
   z^{-1} \\
 \end{bmatrix},
\end{align}
in which $N$ is number of lattice coefficients, and the filter length and order are $2N$ and $2N -1$, respectively. In addition, $\textbf{S}_{m}$ and $\Lambda(z)$ are the biorthogonal lattice and delay matrices with $m = 1,...,N$, respectively. The general form of the biorthogonal lattice matrix $\textbf{S}_{m}$ and of delay matrix $\Lambda(z)$ can be mathematically written as follows:
\begin{equation}
\label{eq:lattice_matrix_delay_matrix_9}
\textbf{S}_{m} = \begin{bmatrix} 1&k_{m} \\k_{m}&1 \\ \end{bmatrix} \quad\text{and}\quad \Lambda(z) = \begin{bmatrix} 1&0 \\0&z^{-1} \\ \end{bmatrix},
\end{equation}
in which $k_{m}$ is the lattice coefficient which determines the biorthogonal lattice matrix $\textbf{S}_{m}$ and can be tuned in the CNN models. In addition, ${T}_{N}(z)$ and ${U}_{N}(z)$ in Eq. \eqref{eq:biorlatticestructure_9} are a mirror image pair (MIP), from which we can later recursive procedure to find the ${H}_{0}(z)$ and ${H}_{1}(z)$ pair of the filter-bank. As ${T}_{N}(z)$ and ${U}_{N}(z)$ are an MIP, we have:
\begin{equation}
\label{eq:TN_and_UN_9}
   {U}_{N}(z) = z^{-(2N-1)}{T}_{N}(z^{-1}). 
\end{equation}
The biorthogonal lattice structure described in  Eq. \ref{eq:biorlatticestructure_9} can be visualized in Fig. \ref{fig:bior-latt-TypeA-AnalysisFilterBank_9}.
Fig. \ref{fig:bior-latt-TypeA-AnalysisFilterBank_9} visualizes the concept demonstrated in Eq. \eqref{eq:biorlatticestructure_9} and Eq. \eqref{eq:lattice_matrix_delay_matrix_9}, showing that biorthogonal filter bank can be written as a multiplication of many biorthogonal matrices determined by their lattice coefficients. Therefore, instead of tuning the filter coefficients directly, the proposed method integrates these lattice coefficients as tunable parameters for the wavelet unit. With lattice coefficients, the filter bank is biorthogonally structured and thus satisfies perfect reconstruction.
\subsection{LS-BiorthLattUwU: Lifting Scheme for Tunable Biorthogonal Wavelet Filters}
Using a lifting scheme, the tunable biorthogonal wavelet unit (LS-BiorUwU) is proposed to relax the constraint of orthogonality to biorthogonality and allows unequal filter lengths in the filter bank of the wavelet unit. In the filter bank structure, the analysis and synthesis parts of the filter bank have the function of decomposing and reconstructing signals, respectively. $H_{0}$ and $H_{1}$ are, correspondingly, low-pass and high-pass filters for the analysis part of the filter bank; whereas $F_{0}$ and $F_{1}$ are, respectively, low-pass and high-pass filters for the synthesis part of the filter bank. With $L$ taps, $H_{0}$ and $H_{1}$ have $\textbf{h}_{0}=[h_0(0), h_0(1),..., h_0(L-1)]$ and $\textbf{h}_{1}=[h_1(0), h_1(1),..., h_1(L-1)]$ as their coefficients, respectively. In orthogonal wavelets, $\textbf{h}_{0}$ and $\textbf{h}_{1}$ are required to have the same length and are related. This requirement can be relaxed if the filter bank is constructed with a lifting scheme from an orthogonal wavelet. Hence, $H_{0}$ and $H_{1}$ can be represented as a matrix multiplication as follows:
\begin{align}
\label{eq:LiftingScheme_9}
 \begin{bmatrix}
   {H}_{0}(z) \\
   {H}_{1}(z) \\
   \end{bmatrix} = \begin{bmatrix}
   \widehat{{H}^{N}_{0}}(z) \\
   \widehat{{H}^{N}_{1}}(z) \\
   \end{bmatrix}= \begin{bmatrix} 1&0 \\P_N(z^2)&1 \\ \end{bmatrix} \begin{bmatrix} 1&0 \\0&z^{-2} \\ \end{bmatrix} \cdots \nonumber\\
  \begin{bmatrix} 1&0 \\P_1(z^2)&1 \\ \end{bmatrix} \begin{bmatrix} 1&0 \\0&z^{-2} \\ \end{bmatrix}
  \begin{bmatrix}
   \widehat{{H}^{0}_{0}}(z) \\
   \widehat{{H}^{0}_{1}}(z) \\
 \end{bmatrix},
\end{align}
in which $\widehat{{H}^{N}_{0}}(z)$ and $\widehat{{H}^{N}_{1}}(z)$ are the final filter pairs constructed after $N$ lifting steps from the $\widehat{{H}^{0}_{0}}(z)$ and $\widehat{{H}^{0}_{1}}(z)$ pair of an orthogonal wavelet. In addition, for $k$ in the range from 1 to $N$, $P_k(z)$ is the lifting step function, which can be represented as follows:
\begin{equation}
\label{eq:LiftingStep_9}
P_k(z) = -a_k + a_kz^{-2k}\mbox{ for }k \mbox{ in }[1, N],
\end{equation}
where $a_k$ is the tunable parameter in the biorthogonal wavelet unit. In addition, the proposed tunable biorthgonal wavelet with lifting scheme can be implemented as the following recursive algorithm:
\begin{align}
\label{eq:recusrive_9}
&\begin{bmatrix}
   \widehat{{H}^{k}_{0}}(z) \\
   \widehat{{H}^{k}_{1}}(z) \\
   \end{bmatrix}= \begin{bmatrix} 1&0 \\P_k(z^2)&1 \\ \end{bmatrix} \begin{bmatrix} 1&0 \\0&z^{-2} \\ \end{bmatrix}\begin{bmatrix}
   \widehat{{H}^{k-1}_{0}}(z) \\
   \widehat{{H}^{k-1}_{1}}(z) \\
   \end{bmatrix}\nonumber\\
   &=\begin{bmatrix}
   \widehat{{H}^{k-1}_{0}}(z) \\
   -a_k\widehat{{H}^{k-1}_{0}}(z)+z^{-2}\widehat{{H}^{k-1}_{1}}(z)+a_kz^{-4k}\widehat{{H}^{k-1}_{0}}(z) \\
   \end{bmatrix},
\end{align}
for $k$ in the range from 1 to $N$. In this work, Haar or Bior1.1 is used for $\widehat{{H}^{0}_{0}}(z)$ and $\widehat{{H}^{0}_{1}}(z)$ initialization. 

\subsection{Implementation in 3D Retina Layer Segmentation Model}
\begin{figure}[t!]
\begin{center}
\includegraphics[width=\linewidth]{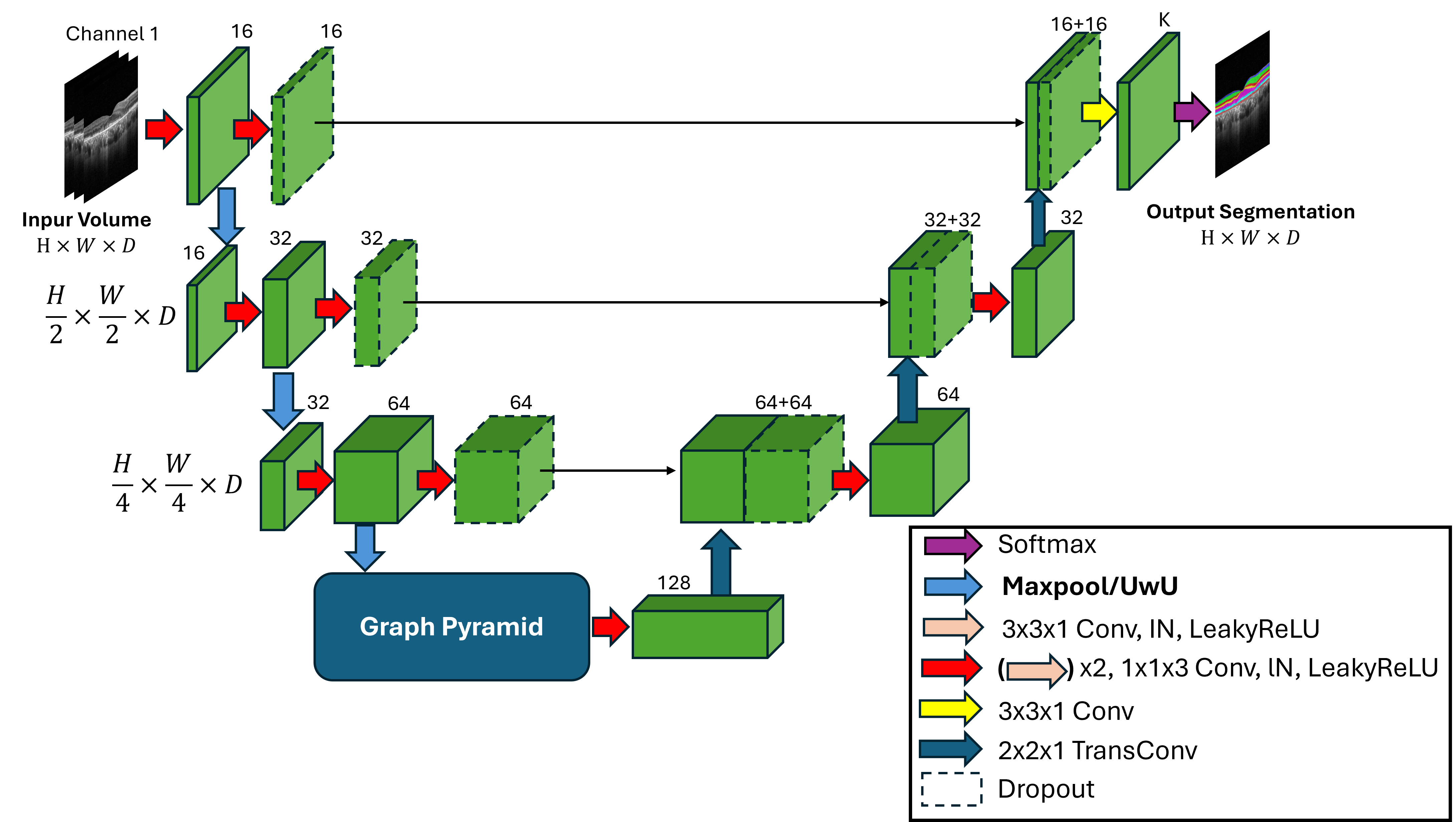}
\end{center}
\caption{\justifying{Implementation of UwU block to replace maxpooling in MGU-Net.}} 
\label{fig:MGU-Net_UwU}
\end{figure}
\begin{figure}[t!]
\begin{center}
\includegraphics[width=0.9\linewidth]{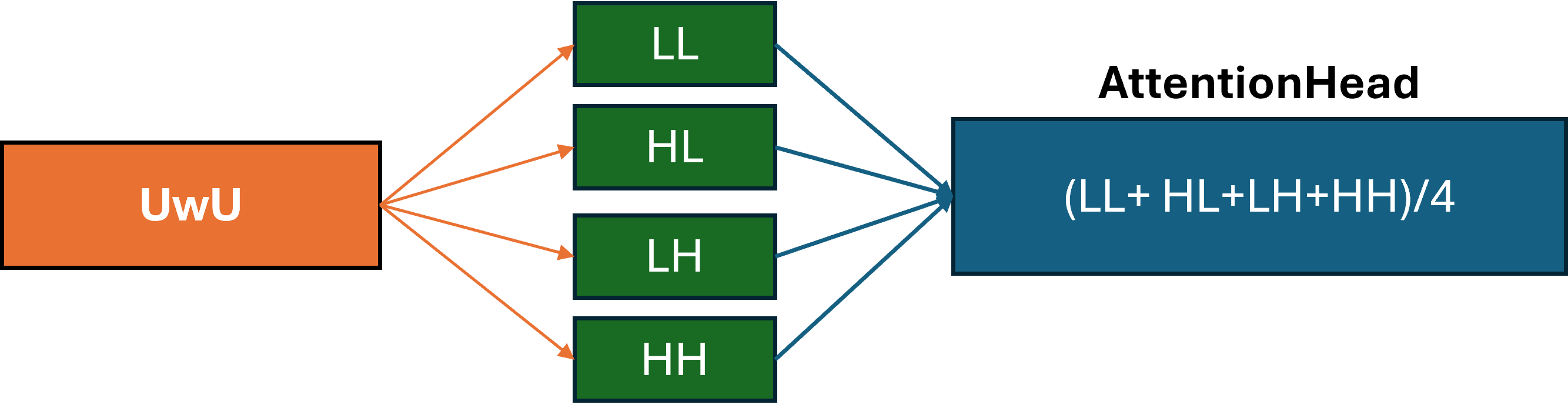}
\end{center}
\caption{\justifying{Features from the UwU subbands (LL, HL, LH, HH) are combined by the attention head.}} 
\label{fig:UwU_heads}
\end{figure}

To enhance spatial detail preservation and frequency-aware learning in retinal OCT segmentation, we replace traditional max-pooling layers in MGU-Net with wavelet-inspired UwU modules—OrthLattUwU, BiorthLattUwU, and LS-BiorthLattUwU, shown in Fig. \ref{fig:MGU-Net_UwU}. These modules perform learnable downsampling using tunable lattice filter banks constrained by stopband energy, enabling better retention of high-frequency and edge information. Additionally, we integrate attention head to adaptively aggregate the subbands (LL, HL, LH, HH), shown in Fig. \ref{fig:UwU_heads}. This replacement allows the modified MGU-Net (UwU-MGUNet) to achieve improved anatomical consistency and segmentation accuracy, particularly in complex pathological regions.
\section{Experiments and Results}
\begin{table*}[!t]
\centering
\scriptsize
\renewcommand{\arraystretch}{0.925}
\setlength{\tabcolsep}{4pt}

\begin{tabular}{|p{3cm}|c|c||c|c|}
\hline
\multicolumn{1}{|c|}{\textbf{MGUNet (Maxpool-Baseline)}} & \multicolumn{2}{c||}{\textbf{Average Dice:} 0.8995} & \multicolumn{2}{c|}{\textbf{Average Accuracy:} 0.9847} \\
\hline
\multicolumn{1}{|c|}{\textbf{UwU (equal filter length)}} & \multicolumn{2}{c||}{\textbf{OrthLattUwU-AttentionHead-MGUNet}} & \multicolumn{2}{c|}{\textbf{BiorthLattUwU-AttentionHead-MGUNet}} \\
\hline
\textbf{} & \textbf{Avg. Dice} & \textbf{Avg. Accuracy} & \textbf{Avg. Dice} & \textbf{Avg. Accuracy} \\
\hline
\multicolumn{1}{|c|}{2-Tap} & 0.9016 (ini-Haar) & 0.9849 (ini-Haar) & 0.9010 (ini-Zeros) & 0.9848 (ini-Zeros) \\
\multicolumn{1}{|c|}{4-Tap} & 0.9003 (ini-DB2) & 0.9847 (ini-DB2) & 0.9029 (ini-Zeros) & 0.9851 (ini-Zeros) \\
\multicolumn{1}{|c|}{6-Tap} & 0.9009 (ini-DB3) & 0.9847 (ini-DB3) & 0.9017 (ini-Zeros) & 0.9849 (ini-Zeros) \\
\multicolumn{1}{|c|}{8-Tap} & 0.8991 (ini-DB4) & 0.9845 (ini-DB4) & 0.8983 (ini-Zeros) & 0.9845 (ini-Zeros) \\
\hline
\multicolumn{1}{|c|}{\textbf{UwU (unequal filter length)}} & \multicolumn{4}{c|}{\textbf{LiftingScheme-BiorthLattUwU-AttentionHead-MGUNet}} \\
\hline
\textbf{} & \multicolumn{2}{c|}{\textbf{Avg. Dice}} & \multicolumn{2}{c|}{\textbf{Avg. Accuracy}} \\
\hline
\multicolumn{1}{|c|}{1-LiftingStep} & \multicolumn{2}{c|}{0.9002} & \multicolumn{2}{c|}{0.9846} \\
\multicolumn{1}{|c|}{2-LiftingSteps} & \multicolumn{2}{c|}{\textbf{0.9030}} & \multicolumn{2}{c|}{\textbf{0.9852}} \\
\multicolumn{1}{|c|}{3-LiftingSteps} & \multicolumn{2}{c|}{0.9015} & \multicolumn{2}{c|}{0.9849} \\
\multicolumn{1}{|c|}{4-LiftingSteps} & \multicolumn{2}{c|}{0.9014} & \multicolumn{2}{c|}{0.9849} \\
\multicolumn{1}{|c|}{5-LiftingSteps} & \multicolumn{2}{c|}{0.9010} & \multicolumn{2}{c|}{0.9848} \\
\multicolumn{1}{|c|}{6-LiftingSteps} & \multicolumn{2}{c|}{0.9029} & \multicolumn{2}{c|}{0.9851} \\
\multicolumn{1}{|c|}{7-LiftingSteps} & \multicolumn{2}{c|}{0.9017} & \multicolumn{2}{c|}{0.9849} \\
\multicolumn{1}{|c|}{8-LiftingSteps} & \multicolumn{2}{c|}{0.9026} & \multicolumn{2}{c|}{0.9850} \\
\hline
\end{tabular}
\caption{Comparison of Segmentation Performance of UwU Variants with Attention Head integrated in MGUNet.}
\label{table:UwU_MGUNet_AttentionHead}
\end{table*}
\begin{figure*}[!t]
\begin{center}
\includegraphics[width=0.92\linewidth]{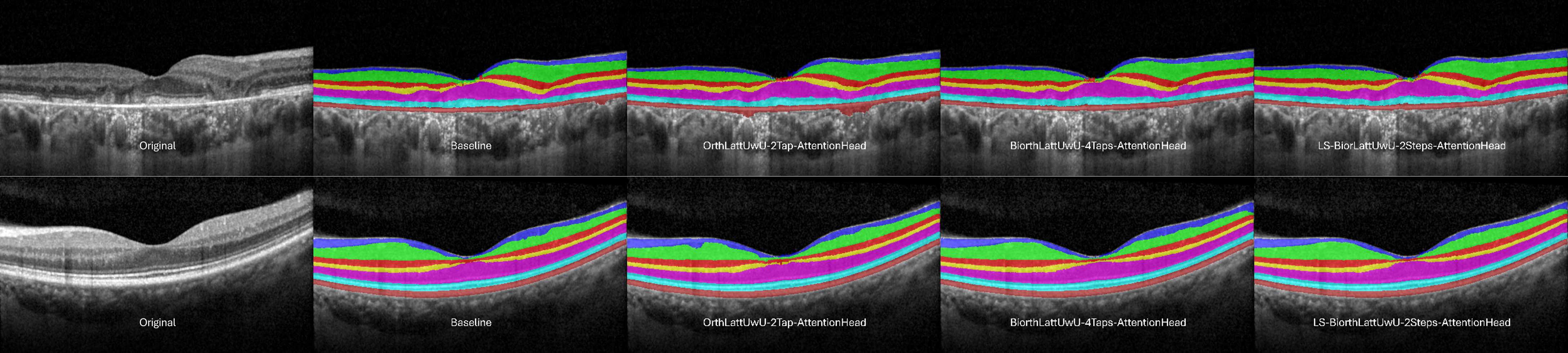}
\end{center}
\caption{\justifying{Segmentation results on two OCT volume samples—one from a diseased retina (top) and one from a normal subject (bottom). The figure compares outputs from the MGUNet baseline and three variants with wavelet-based pooling: OrthLattUwU-2Tap, BiorthLattUwU-4Tap, and LS-BiorthLattUwU-2Step.}} 
\label{fig:SegmentVis}
\end{figure*}
This study was conducted in accordance with the Declaration of Helsinki and HIPAA regulations. Written informed consent was obtained from all participants undergoing surgery, and all patient data were anonymized prior to analysis. The JRC dataset \cite{Yiqian3} was used to evaluate the proposed method against clinically available OCT segmentation solutions for retinal layer delineation across a range of ocular pathologies. The dataset comprises 160 horizontal and vertical OCT volumes acquired using the Heidelberg Spectralis system \cite{heidelberge}. Of these, 30 volumes were manually corrected for eight retinal layers using the Heidelberg HEYEX software, based on the system’s initial automated segmentation. Annotations were performed by six trained retinal MD fellows at the Jacobs Retina Center and subsequently reviewed and refined by an experienced grader. The remaining volumes—those without manual correction—were split into 142 for training and 18 for validation, using Heidelberg’s automated segmentation as ground truth. Each OCT volume has a resolution of 496 × 512 × 49 voxels, corresponding to a physical size of 1.9 × 5.8 × 5.8 mm\textsuperscript{3}. The dataset includes both normal cases and patients diagnosed with wet and dry age-related macular degeneration (AMD), nonproliferative diabetic retinopathy (NPDR), epiretinal membrane (ERM), central retinal vein occlusion (CRVO), retinal detachment, macular hole, chorioretinopathy, and other retinal conditions. Pathology labels and diagnostic information were recorded by ophthalmologists at the Jacobs Retina Center.
To evaluate the effectiveness of wavelet-based downsampling in retinal OCT segmentation, we integrated OrthLatt-UwU, BiorthLatt-UwU, and LS-BiorthLatt-UwU modules into a 3D segmentation pipeline based on the MGU-Net architecture. These modules replaced conventional maxpooling layers and were tested on the JRC retinal OCT datase. Segmentation performance was assessed quantitatively using Dice scores and accuracy. The OrthLatt-UwU implementation, using a 2-tap orthogonal lattice structure, showed a clear improvement over the baseline maxpooling method. The BiorthLatt-UwU further enhanced performance by leveraging the flexibility of biorthogonal filter banks. The most significant gain, however, was observed with the LS-BiorthLatt-UwU variant with an attention head shown in Table \ref{table:UwU_MGUNet_AttentionHead}. This configuration achieved the highest Dice scores and accuracy among all tested models. These results support our hypothesis that longer high-pass filters, as used in LS-BiorthLatt-UwU, are more effective in capturing and preserving fine-grained anatomical features in OCT volumes.
These results collectively demonstrate that LS-BiorthLatt-UwU, especially when combined with attention-based subband fusion, offers a powerful alternative to traditional pooling layers in 3D medical image segmentation—capturing clinically relevant details that are often lost in standard CNN architectures. The segmentation of the examined models are shown in Fig. \ref{fig:SegmentVis}

\section{Conclusion}
\label{sec:conclude}
This study is the first to demonstrate the effectiveness of tunable wavelet-based pooling modules for 3D retinal OCT segmentation. By replacing conventional pooling layers with OrthLattUwU, BiorthLattUwU, and LS-BiorthLattUwU, we show that learnable wavelet filters can better preserve anatomical detail and spatial coherence. Among the variants, LS-BiorthLattUwU achieves the best performance, validating the potential of lifting-based biorthogonal designs. Our results establish a new direction for incorporating wavelet theory into 3D medical image analysis, with promising implications for high-precision retinal diagnostics.
\bibliographystyle{ieeetr}
\bibliography{refs}

\end{document}